\pdfoutput=1

\documentclass[11pt]{article}

\usepackage[]{acl}

\usepackage{times}
\usepackage{latexsym}

\usepackage{devanagari}
\usepackage{graphicx}
\usepackage[table]{colortbl}
\usepackage{booktabs}

\usepackage[T1]{fontenc}

\usepackage[utf8]{inputenc}

\usepackage{microtype}

\title{Don't Overlook the Grammatical Gender: Bias Evaluation for Hindi-English Machine Translation}

\author{Pushpdeep Singh \\
  National Institute of Technology, Hamirpur \\ Anu, Hamirpur, India \\
  \texttt{pushpdeep30@gmail.com} \\}

\begin{document}
\maketitle
\begin{abstract}
Neural Machine Translation (NMT) models, though state-of-the-art for translation, often reflect social biases, particularly gender bias. Existing evaluation benchmarks primarily focus on English as the source language of translation. For source languages other than English, studies often employ gender-neutral sentences for bias evaluation, whereas real-world sentences frequently contain gender information in different forms. Therefore, it makes more sense to evaluate for bias using such source sentences to determine if NMT models can discern gender from the grammatical gender cues rather than relying on biased associations. To illustrate this, we create two gender-specific sentence sets in Hindi to automatically evaluate gender bias in various Hindi-English (HI-EN) NMT systems. We emphasise the significance of tailoring bias evaluation testsets to account for grammatical gender markers in the source language. 

\end{abstract}

\section{Introduction}

State-of-the-art NMT models develop gender bias \citep{prates2019assessing} during their training by picking up spurious correlations in their training data. Due to their poor coreference resolution ability, they often rely on biased associations with, say, occupation terms to disambiguate the gender of pronouns (\citealp{stanovsky-etal-2019-evaluating, saunders-etal-2020-neural}). Also, in many cases, these models translate gender-neutral sentences into gendered ones (\citealp{prates2019assessing, cho-etal-2019-measuring}) especially giving a `masculine default' translation or one which is based on a stereotype. 
This problem also exists for HI-EN Machine Translation \citep{ramesh-etal-2021-evaluating}. When put to use, such systems can cause various harms \citep{savoldi-etal-2021-gender}. Thus, evaluating and mitigating such biases from NMT models is critical to ensure fairness. 

Prior research evaluating gender bias in machine translation has predominantly centered around English as the source language (\citealp{stanovsky-etal-2019-evaluating, bentivogli-etal-2020-gender, costa-jussa-etal-2020-gebiotoolkit, currey-etal-2022-mt, rarrick2023gate}). However, these evaluation methods or testsets don't seamlessly extend to other source languages, especially the ones with grammatical gender. For instance, in Hindi, elements like pronouns, adjectives, and verbs are inflected with gender. Nonetheless, prior studies in other source languages often utilise gender-neutral sentences for bias evaluation using TGBI score (\citealp{cho-etal-2019-measuring, ramesh-etal-2021-evaluating}). In our experiments with TGBI Evaluation, we found that it doesn't expose gender bias present in various NMT models.

Therefore, in this work, we propose to evaluate NMT models for bias using sentences with grammatical gender cues of the source language. This allows us to ascertain whether NMT models can discern the accurate gender from context or if they depend on biased correlations. Using Hindi as the source language in NMT, we develop two bias evaluation benchmarks: \emph{OTSC-Hindi} and \emph{WinoMT-Hindi} and evaluate various HI-EN NMT models for gender bias. We highlight the importance of creating such benchmarks for source languages with expressive gender markers.

\begin{table*}[tbh!]
  \centering
  \begin{tabular}{@{}lcccccccc@{}}
    \toprule
    & \multicolumn{2}{c}{\textbf{IT}}                                                                & \multicolumn{2}{c}{\textbf{GT}}                                                                & \multicolumn{2}{c}{\textbf{MS}}                                                                 & \multicolumn{2}{c}{\textbf{AWS}}                                                             \\
\multicolumn{1}{c}{\textbf{Sentence Set}}    & \multicolumn{1}{c}{$p_m$} & \multicolumn{1}{c}{$p_f$} & \multicolumn{1}{c}{$p_m$} & \multicolumn{1}{c}{$p_f$} & \multicolumn{1}{c}{$p_m$} & \multicolumn{1}{c}{$p_f$}  & \multicolumn{1}{c}{$p_m$} & \multicolumn{1}{c}{$p_f$} \\ \midrule
    \multicolumn{1}{l|}{\textit{\small{Female Speaker, Female Friend}}} & \textbf{98.41}                                                  & \multicolumn{1}{c|}{1.59$^*$}      & 1.68                         & \multicolumn{1}{c|}{\textbf{98.32$^*$}}      & \textbf{98.97}                                   & \multicolumn{1}{c|}{1.03$^*$}      & \textbf{95.61}                                                    & 4.39$^*$                           \\
    \multicolumn{1}{l|}{\textit{\small{Female Speaker, Male Friend}}} & {\textbf{99.25$^*$}}                                  & \multicolumn{1}{c|}{0.75}      & \textbf{90.66$^*$}  & \multicolumn{1}{c|}{9.34}      & \textbf{99.72$^*$}                                                     & \multicolumn{1}{c|}{0.28}      & \textbf{95.70$^*$}                                              & 4.30                            \\
    \multicolumn{1}{l|}{\textit{\small{Male Speaker, Female Friend}}} & \textbf{99.35}                                                   & \multicolumn{1}{c|}{0.65$^*$}      & 2.43                                & \multicolumn{1}{c|}{\textbf{97.57$^*$}}      & \textbf{66.01}                                       & \multicolumn{1}{c|}{33.99$^*$}      & \textbf{99.29}                                                    & 2.71$^*$                            \\
        \multicolumn{1}{l|}{\textit{\small{Male Speaker, Male Friend}}} & \textbf{99.91$^*$}                                                      & \multicolumn{1}{c|}{0.09}      & \textbf{96.45$^*$}                               & \multicolumn{1}{c|}{3.55}      & \textbf{98.60$^*$}                                          & \multicolumn{1}{c|}{1.40}      & \textbf{97.48$^*$}                                               & 2.52                            \\\bottomrule
    \end{tabular}%

  \caption{
    Evaluation using the OTSC-Hindi test set. Here, $p_m$ and $p_f$ are the percentages of sentences translated as male and female for the speaker’s friend. $*$ corresponds to the percentage of sentences translated into the true label for each sentence set. Bold values indicate the maximum percentage of sentences translated into a single gender class.}
    \label{tab:results_otsc}
\end{table*}
\begin{table}[tbh!]
  \centering
  \begin{tabular}{@{}lcccc@{}}
    \toprule
    & \multicolumn{1}{c}{\textbf{$Acc$}} & \multicolumn{1}{c}{\textbf{$\Delta_G$}} & \multicolumn{1}{c}{\textbf{$\Delta_S$}} & \multicolumn{1}{c}{\textbf{$N$}} \\ \midrule
    \textit{\small{\textbf{IndicTrans}}} &48.9 &48.5 &-0.1$^{\bullet}$ &6.2    \\
    \textit{\small{\textbf{Google Translate}}} &69.0$^{\star}$ &10.6$^{\diamond}$ &-3.8 &5.3     \\ 
    \textit{\small{\textbf{Microsoft Translator}}} &57.7 &32.9 &0.2$^{\bullet}$ &4.1     \\ 
    \textit{\small{\textbf{AWS Translate}}} &49.9 &51.9 &-0.2$^{\bullet}$ &2.8     \\ 
    \bottomrule
    \end{tabular}%
  \caption{Comparison of performance of various NMT Models on WinoMT-Hindi. ${\star}$ indicates significantly highest value, ${\diamond}$ indicates significantly lowest value, ${\bullet}$ indicates near about values for \emph Acc, $\Delta_G$ and $\Delta_S$, respectively. \emph{N} represents percentage of sentences translated into gender-neutral form. }
  \label{tab:results-winomt}
  \end{table}

\section{Experimental Setup}

We create two bias evaluation testsets in Hindi to evaluate gender bias: Occupation Testset with Simple Context (\textit{OTSC-Hindi}) and \textit{WinoMT-Hindi}, which account for diverse grammatical gender cues in Hindi. Using these, we evaluate four HI-EN NMT Models: IndicTrans (IT) \citep{ramesh-etal-2022-samanantar}, Google Translate (GT), Microsoft Translator (MS) and AWS Translate (AWS). Previous work by \citet{ramesh-etal-2021-evaluating} focused only on the gender-neutral side of Hindi; however, we demonstrate how we can account for gender-specified context for better evaluation of bias. Code and data are publicly available\footnote{\url{https://github.com/iampushpdeep/Gender-Bias-Hi-En-Eval}}.

\section{OTSC-Hindi}

Similar to \citet{escude-font-costa-jussa-2019-equalizing}, we create a Hindi version with grammatical gender cues: `` {\dn m\4{\qva} us\? kAPF smy s\?} \{{\dn jAntA{\rs ,\re}jAntF}\} {\dn \8{h}\1{\rs,\re}}\{{\dn m\?rA{\rs,\re}m\?rF}\} {\dn do-t} \textbf{[occupation]} {\dn kA kAm} \{{\dn krtA{\rs ,\re}krtF}\} {\dn h\4.} ''(\textit{I have known [him/her] for a long time, my friend works as a [occupation].}) Unlike the English version, this template specifies the gender of the speaker using a gender-inflected verb, i.e. ``{\dn jAntA}(\textit{jaanta})''(m) and ``{\dn jAntF}(\textit{jaanti})''(f). The possessive pronoun is also gender inflected based on the gender of the speaker's friend: ``{\dn m\?rA}(\textit{mera})''(m) and ``{\dn m\?rF}(\textit{meri})''(f) which decides the use of verb ``{\dn krtA}(\textit{karta})''(m) and ``{\dn krtF}(\textit{karti})'' (f).  However, the English translation only specifies the gender of speaker's friend. Based on four possibilities in the template, we construct these four sets of sentences.

\section{WinoMT-Hindi}
\citet{stanovsky-etal-2019-evaluating} composed a challenge set called \textit{WinoMT} for evaluating gender bias in NMT models. We contextualize this test set for the evaluation of bias in HI-EN Translation by manually creating ``WinoMT-Hindi'', which consists of 704 WinoBias-like sentences \cite{zhao-etal-2018-gender} in Hindi but modified to include gender cues of the language, mainly: gender-inflected adjectives, postpositions, and verbs. We can mark the gender of the target (English) by simply checking for the presence of male pronouns (he, him or his) or female pronouns (she or her) in the translation. Interestingly, we also observe that a few sentences are translated into a gender-neutral form. For gender bias evaluation, we use the metrics \emph{Acc}, $\Delta_G$ and $\Delta_S$ given by \citet{stanovsky-etal-2019-evaluating}.

\section{Evaluation Results}
The results are shown in Table~\ref{tab:results_otsc} and Table~\ref{tab:results-winomt}. Based on these results, most models show heavy bias against the female gender and prefer masculine default in translations. On WinoMT-Hindi, these models perform only as good as random guessing, while Google Translate performs better at discerning gender using the grammatical gender cues in the source. On the other hand, evaluation based on TGBI (results not shown here) gives almost similar performance on all NMT systems which shows that it is not good at exposing gender bias.

\section{Conclusion and Future Work}
We highlighted the need for contextualising NMT bias evaluation for non-English source languages. In future, we plan to extend our evaluation beyond following a particular template and be accommodating to all gender identities \citep{10.1145/3531146.3534627}. 

\bibliography{anthology,custom}

\end{document}